# MergeSAM: UNSUPERVISED CHANGE DETECTION OF REMOTE SENSING IMAGES BASED ON THE SEGMENT ANYTHING MODEL

*Meiqi Hu[1], Lingzhi Lu[2], Chengxi Han[3], Xiaoping Liu[1]*

[1]School of Geography and Planning, Sun Yat-sen University, Guangzhou 510275, China
[2] Unit 33269 of the PLA, Lanzhou 730000，China
[3]Intelligent Science & Technology Academy Limited of CASIC, Beijing 100043, China

## ABSTRACT

Recently, large foundation models trained on vast datasets have demonstrated exceptional capabilities in feature extraction and general feature representation. The ongoing advancements in deep learning-driven large models have shown great promise in accelerating unsupervised change detection methods, thereby enhancing the practical applicability of change detection technologies. Building on this progress, this paper introduces MergeSAM, an innovative unsupervised change detection method for high-resolution remote sensing imagery, based on the Segment Anything Model (SAM). Two novel strategies, MaskMatching and MaskSplitting, are designed to address real-world complexities such as object splitting, merging, and other intricate changes. The proposed method fully leverages SAM's object segmentation capabilities to construct multitemporal masks that capture complex changes, embedding the spatial structure of land cover into the change detection process.

## 1. INTRODUCTION

With the rapid advancement of remote sensing technology, high-resolution satellite imagery has emerged as a crucial data source supporting applications in Earth observation, environmental monitoring, and urban planning [1], [2]. As an essential method for analyzing differences between images from different time periods, change detection has become one of the core tasks in remote sensing image processing [3], [4].

Traditional change detection methods usually rely on prominent features or statistical characteristics, failing to adapt to complex scenes and data diversity, which limits their practical effectiveness [5]. Recently, the quick development of deep learning has greatly advanced change detection in remote sensing imagery [6]. Deep learning techniques enable models to automatically extract complex feature representations from vast amounts of data, thereby enhancing the accuracy and robustness of change detection [7]. However, most deep learning-based change detection methods rely on extensive manual labeling, incurring high costs and limiting scalability in real-world applications.

By contrast, unsupervised change detection methods offer significant advantages over supervised methods by eliminating the need for labeled data, making it a promising solution to the limited labeling issue in remote sensing. The rise of large pre-trained models, with their powerful image feature extraction and general representation capabilities, has opened new opportunities for unsupervised change detection. Visual foundation models [8], exemplified by the Segment Anything Model (SAM) [9], have achieved remarkable success in natural image segmentation. SAM stands out for its powerful segmentation capabilities, zero-shot generalization, and promptability, offering exceptional flexibility and adaptability across a wide range of vision tasks [10]. Moreover, recent research has focused on developing parameter-efficient fine-tuning methods to enhance the SAM model's ability to learn domain-specific features, showing great potential in remote sensing classification [11], object detection [12], and change detection [13]. Zheng et al., proposed a pioneering unsupervised change detection method, named AnyChange [14]. Concretely, a bitemporal latent matching strategy is designed to leverage segmentation masks and encoder features for instance-level change deatection, validated on multiple remote sensing datasets. However, AnyChange faces limitations when handling complex changes, such as object splitting and merging. Therefore, how to harness SAM's powerful zero-shot transfer capability to unlock its potential for unsupervised change detection in remote sensing, remains a promising area for further investigation [15], [16].

Inspired by this, a novel unsupervised remote sensing change detection strategy, Bitemporal Mask Matching, is proposed. This method fully utilizes the object segmentation capabilities of the SAM to construct multitemporal masks that capture complex changes, such as object splitting and merging, while embedding spatial structure information of land cover into the change detection process. The resulting model, MergeSAM, is highly effective in addressing complex change detection in remote sensing imagery.

## 2. METHODOLOGY

### 2.1 Segment Anything Model

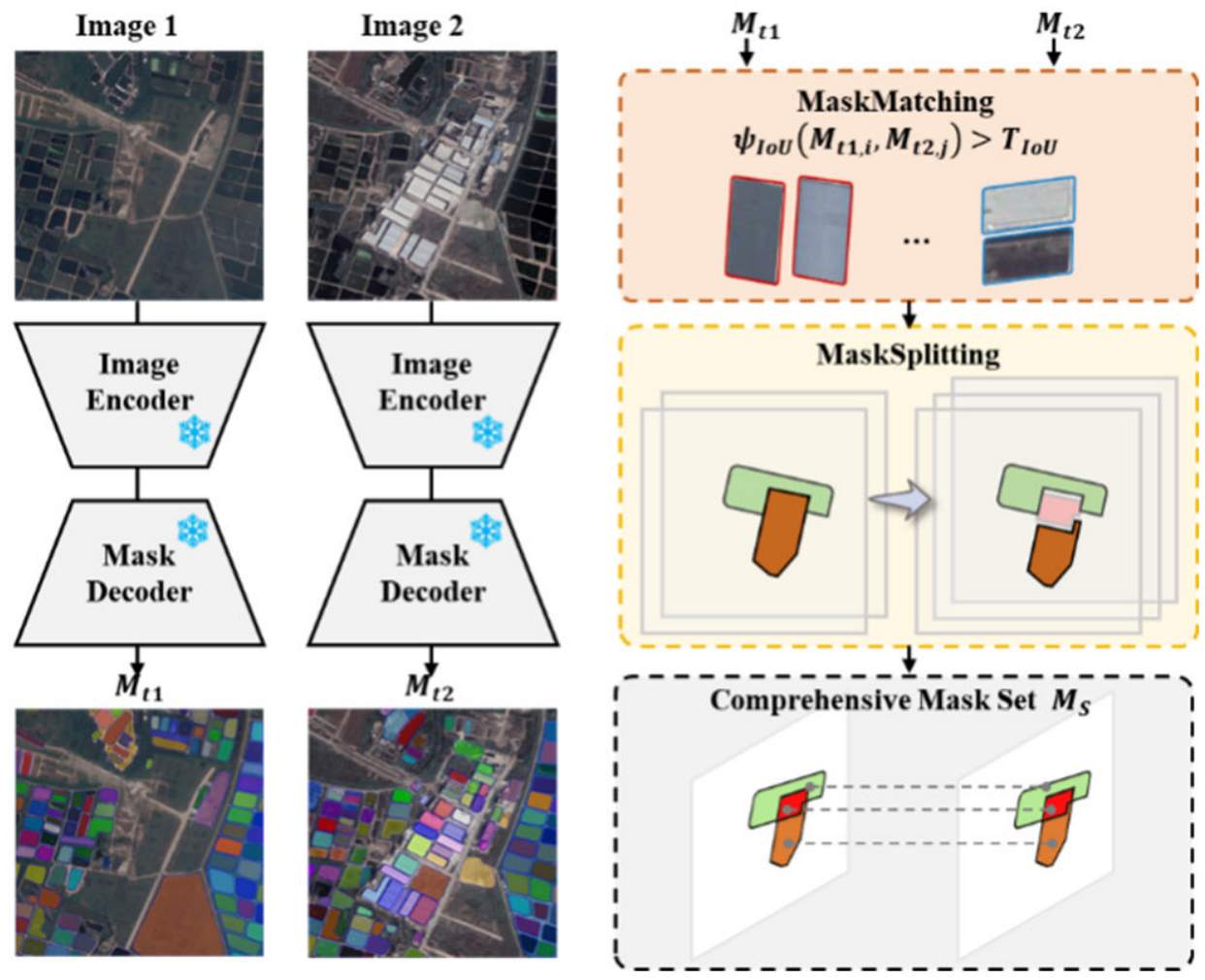

**Fig. 1 The structure of the proposed MergeSAM**

The Segment Anything Model (SAM) is a highly advanced vision segmentation model that exhibits exceptional zero-shot generalization capabilities and notable promptability. It is capable of segmenting arbitrary objects within an image by leveraging various forms of user-defined prompts, such as points, boxes, or masks. SAM requires no task-specific retraining, allowing it to seamlessly adapt to new images and perform effectively across a wide range of segmentation tasks.

SAM consists of three primary components: the image encoder, the prompt encoder, and the mask decoder. The Image Encoder extracts both global and local features from the input image, transforming it into a high-dimensional representation suitable for further processing. The Prompt Encoder encodes the segmentation cues provided by the user (in the form of points, boxes, or masks), guiding the model's focus during segmentation. The Mask Decoder then generates precise object masks based on these encoded inputs. In the absence of explicit prompts, SAM automatically generates a uniform grid of points across the entire image as an implicit prompt, facilitating the extraction of masks for all objects within the scene. This innovative design allows SAM to perform efficient and accurate segmentation without the need for additional training or specific input, ensuring robust performance across diverse segmentation tasks.

### 2.2 MergeSAM

Fig. 1 illustrates the structure of the MergeSAM model. Bitemporal Mask Matching primarily integrates the spatial structural information of object masks from different images to reflect the changes of ground objects. Additionally, the feature differences between multitemporal masks characterize the complex transformation of objects over time.

Concretely, the two strategies, MaskSplitting and MaskMatching, are designed to effectively capture the distribution of object changes across multitemporal images. MaskSplitting constructs multitemporal masks, which aim at efficiently addressing the splitting and merging of complex objects transformation, such as the transformation of large barren lands into urban areas during city expansion. The prevalent registration errors or variations in solar angles may lead to inconsistencies in the segmentation boundaries of the same object. To address this, MaskMatching is tailored to align masks corresponding to the same object across multitemporal images.

**MaskMatching Strategy.** Given two remote sensing images, $x_1$ and $x_2$, the SAM model automatically generates segmentation masks $M_{t1}$ and $M_{t2}$ for each image. Simultaneously, the image encoder extracts feature embeddings $e_1$ and $e_1$ for $x_1$ and $x_2$. The MaskMatching strategy selects object pairs based on the intersection-over-union (IoU) between the corresponding spatially aligned object masks from the two images, prioritizing those with a large intersection. The IoU threshold, denoted as $T_{IoU}$, is set to a fixed value, enabling unsupervised matching of multitemporal object sets with slight displacement or deformation differences.

**MaskSplitting Strategy.** The MaskSplitting strategy acknowledges that an object may not always change as a whole; instead, part of the object may change while the remaining part stays the same. Firstly, the matching masks are removed from the segmentation masks, leaving behind the potentially changed masks from each image, denoted as $M'_{t1}$ and $M'_{t2}$. Then the MaskSplitting strategy splits the masks $M'_{t1}$ and $M'_{t2}$ that have intersections, which forms a multitemporal mask set $M$. The intersecting portions of multitemporal objects may reflect the ground objects' transformation over time. The matching masks and multitemporal masks together constitute the comprehensive mask set $M_S$.

With the MaskMatching and MaskSplitting strategies, multitemporal homogeneous regions of ground objects are extracted from the two temporal images. Each multitemporal mask acts as an analysis unit, where the average feature within the mask serves as the feature embedding and Mean Squared Error (MSE) as the similarity measurement. Final results are obtained through thresholding using the Otsu algorithm. MergeSAM integrates the spatial structure of multitemporal objects, effectively accounting for real-world phenomena such as splitting, merging, and complex changes, thereby improving the performance of unsupervised change detection.

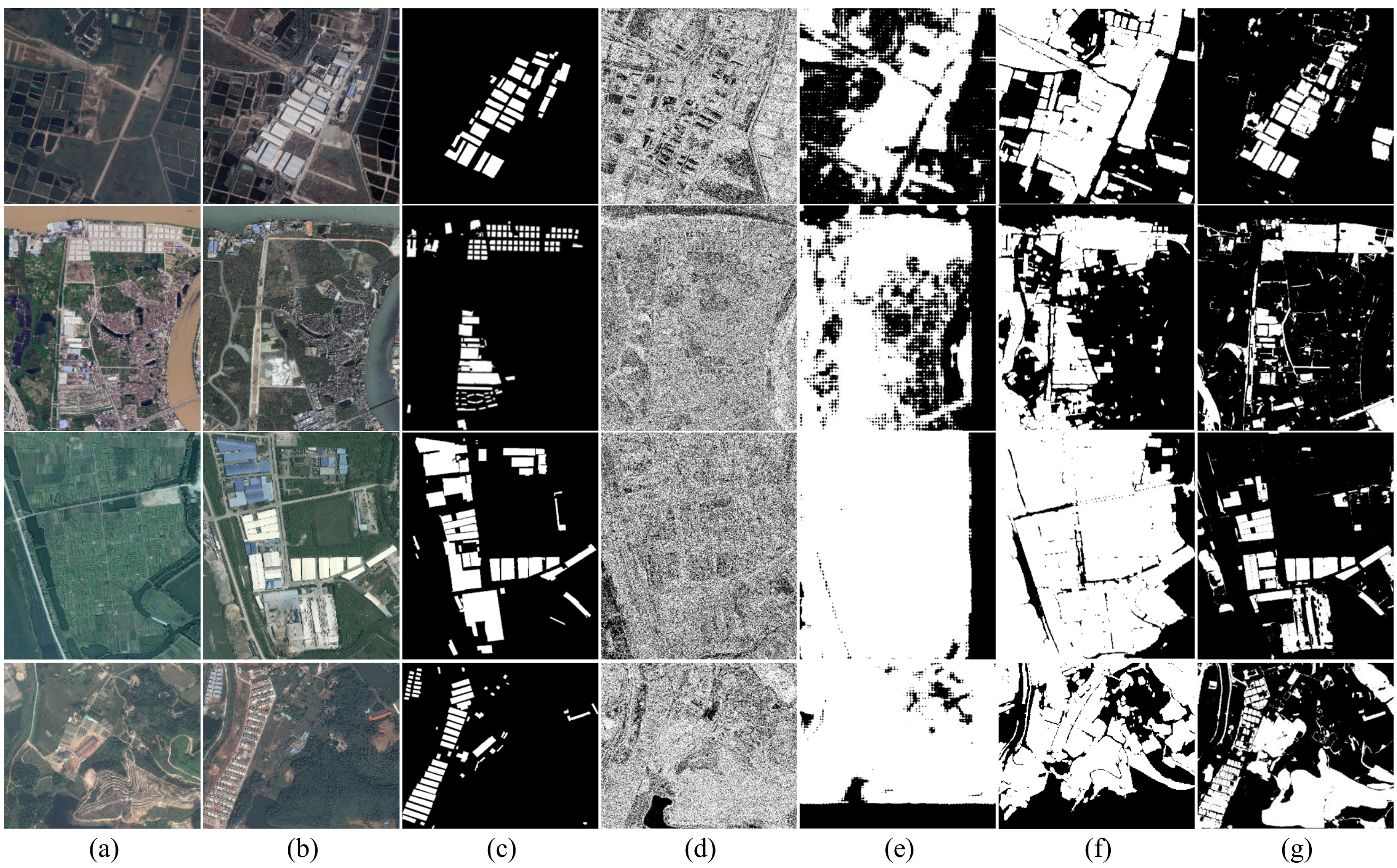

**Fig. 2** The change detection results of GZ_CD_data dataset. Pseudo visualization of (a) image 1 and (b) image 2. (c) Reference, (d) CVA, (e) CVA+SAM, (f) AnyChange, (g) MergeSAM.

## 3. EXPERIMENTS

### 3.1 Experimental Setting

**Dataset Description.** The GZ_CD_data[17] is a high-resolution remote sensing change detection dataset collected from 2006 to 2019, covering the suburbs of Guangzhou, China. The dataset consists of 20 pairs of Very High Resolution (VHR) images, each capturing seasonal variations. The spatial resolution of the images is 0.55 meters per pixel, with image sizes ranging from 1,006 × 1,168 pixels to 4,936 × 5,224 pixels.

**Parameter Setting.** For object proposal generation, we adopt a point per side of 64, an NMS threshold of 0.7, a predicted IoU threshold of 0.5, and a stability score threshold is 0.8. For MaskMatching Strategy, $T_{IoU}$=0.75. Given the large size of the image, the image is scaled proportionally, with the longest side resized to 1600.

**Metrics.** Five Commonly used evaluation metrics for change detection include Precision (Prec.), Recall (Rec.), F1, OA, and Kappa. F1 and Kappa are composite metrics, where higher values indicate better overall performance in change detection. Recall measures the proportion of true changes correctly identified, while Precision reflects the proportion of detected changes that are truly correct

**GPU Configuration.** The experiment was conducted using an NVIDIA 2080 GPU.

**Comparative algorithms.** All comparison algorithms are unsupervised remote sensing change detection methods, including classic Change Vector Analysis (CVA) [18], and models based on SAM, CVA+SAM and AnyChange [14] included.

### 3.2 Results and Analysis

The unsupervised change detection maps on GZ_CD_data dataset are presented in Fig. 2. TABLE 1 presents the quantitative evaluation results, with the optimal values highlighted in bold. According to Fig. 2, significant noise appears in its change detection map of CVA. CVA+SAM correctly detects most of the chan ge areas, but it suffers from high false alarm rate, which corresponds to its high recall and low precision on the TABLE 1. In addition, AnyChange successfully identifies changes such as newly constructed buildings and transformation of building updates. However, due to its direct combination of masks of each image for change detection, there are lots of false detections in the results. By contrast, the proposed MergeSAM method detects most of the change areas with fewer false detections and less missed detections. This improvement stems from the MaskMatching and MaskSplitting strategies, which consider complex changes such as object splitting and merging, thus yielding optimal unsupervised change detection performance. As shown in TABLE 1, MergeSAM consistently achieves the best comprehensive change detection results across various SAM backbone configurations. Compared with AnyChange, MergeSAM exhibits a 7% improvement in the F1 score, enhancing precision while maintaining a high recall.

**TABLE 1** The Quantitative Assessment On GZ_CD_data

| Method | Backbone | F1 | Prec. | Rec. | OA | Kappa |
|---|---|---|---|---|---|---|
| CVA | - | 16.30 | 9.32 | 64.67 | 40.61 | 0.79 |
| CVA+SAM | Vit-B | 22.54 | 13.00 | **84.82** | 47.89 | 8.33 |
| AnyChange | Vit-B | 24.84 | 14.70 | 79.98 | 56.73 | 11.47 |
| MergeSAM | Vit-B | **31.65** | **20.80** | 66.18 | **74.45** | **20.89** |
| CVA+SAM | Vit-L | 22.25 | 12.83 | **83.72** | 47.70 | 7.99 |
| AnyChange | Vit-L | 25.29 | 15.10 | 77.76 | 58.92 | 12.13 |
| MergeSAM | Vit-L | **31.59** | **21.01** | 63.64 | **75.36** | **20.96** |
| CVA+SAM | Vit-H | 22.29 | 12.87 | **83.03** | 48.25 | 8.06 |
| AnyChange | Vit-H | 25.83 | 15.52 | 77.02 | 60.46 | 12.86 |
| MergeSAM | Vit-H | **30.13** | **20.37** | 57.81 | **76.03** | **19.48** |

## 4. CONCLUSION

To conclude, this paper focuses on how to effectively utilize the powerful feature extraction capabilities of large models in the era of deep learning to promote the development of unsupervised change detection of remote sensing images. A novel unsupervised change detection method named MergeSAM is proposed for high-resolution remote sensing images. Combined with SAM, MergeSAM enables automatic extraction of complex changes without the need for training samples. The effectiveness of the proposed method was tested on a comprehsenive high-resolution remote sensing change detection dataset, and the results demonstrate that the proposed MaskMatching and MaskSplitting strategies can effectively detect fine change parcels with few false detections and missed detections.

## ACKNOWLEDGMENT

This work was supported in part by the China Postdoctoral Science Foundation General Funding Project (2024M763747).

## REFERENCES

[1] M. Burke, A. Driscoll, D. B. Lobell, and S. Ermon, "Using satellite imagery to understand and promote sustainable development," *Science*, vol. 371, no. 6535, p. eabe8628, Mar. 2021.

[2] I. Demir *et al.*, "DeepGlobe 2018: A Challenge to Parse the Earth through Satellite Images," May 17, 2018, *arXiv*: arXiv:1805.06561. doi: 10.48550/arXiv.1805.06561.

[3] T. Bai *et al.*, "Deep learning for change detection in remote sensing: a review," *Geo-spatial Information Science*, vol. 0, no. 0, pp. 1–27, Jul. 2022, doi: 10.1080/10095020.2022.2085633.

[4] D. Wen *et al.*, "Change Detection From Very-High-Spatial-Resolution Optical Remote Sensing Images: Methods, applications, and future directions," *IEEE Geoscience and Remote Sensing Magazine*, vol. 9, no. 4, pp. 68–101, Dec. 2021.

[5] S. Ashbindu, "Review Article Digital change detection techniques using remotely-sensed data," *Int. J. Remote Sens.*, vol. 10, no. 6, 1989.

[6] L. Zhang, L. Zhang, and B. Du, "Deep learning for remote sensing data: a technical tutorial on the state of the art," *IEEE Geosci. Remote Sens. Mag.*, vol. 4, no. 2, pp. 22–40, Jun. 2016, doi: 10.1109/MGRS.2016.2540798.

[7] T. Zhao *et al.*, "Artificial intelligence for geoscience: Progress, challenges, and perspectives," *The Innovation*, vol. 5, no. 5, p. 100691, 2024, doi: 10.1016/j.xinn.2024.100691.

[8] D. Wang *et al.*, "HyperSIGMA: Hyperspectral Intelligence Comprehension Foundation Model," *arXiv preprint arXiv:2406.11519*, 2024.

[9] A. Kirillov *et al.*, "Segment anything," in *Proceedings of the IEEE/CVF International Conference on Computer Vision*, 2023, pp. 4015–4026.

[10] H. Chen, J. Song, and N. Yokoya, "Change Detection Between Optical Remote Sensing Imagery and Map Data via Segment Anything Model (SAM)," Jan. 17, 2024, *arXiv*: arXiv:2401.09019. Accessed: Jan. 22, 2024. [Online]. Available: http://arxiv.org/abs/2401.09019

[11] X. Zhou *et al.*, "MeSAM: Multiscale Enhanced Segment Anything Model for Optical Remote Sensing Images," *IEEE Transactions on Geoscience and Remote Sensing*, pp. 1–1, 2024, doi: 10.1109/TGRS.2024.3398038.

[12] K. Chen *et al.*, "RSPrompter: Learning to Prompt for Remote Sensing Instance Segmentation Based on Visual Foundation Model," *IEEE Trans. Geosci. Remote Sensing*, vol. 62, pp. 1–17, 2024, doi: 10.1109/TGRS.2024.3356074.

[13] L. Ding, K. Zhu, D. Peng, H. Tang, K. Yang, and L. Bruzzone, "Adapting Segment Anything Model for Change Detection in VHR Remote Sensing Images," *IEEE Transactions on Geoscience and Remote Sensing*, vol. 62, pp. 1–11, 2024, doi: 10.1109/TGRS.2024.3368168.

[14] Z. Zheng, Y. Zhong, L. Zhang, and S. Ermon, "Segment Any Change," Feb. 14, 2024, *arXiv*: arXiv:2402.01188. Accessed: May 20, 2024. [Online]. Available: http://arxiv.org/abs/2402.01188

[15] C. Zhang *et al.*, "A Comprehensive Survey on Segment Anything Model for Vision and Beyond," May 19, 2023, *arXiv*: arXiv:2305.08196. doi: 10.48550/arXiv.2305.08196.

[16] D. Li, M. Wang, H. Guo, and W. Jin, "On China's earth observation system: mission, vision and application," *Geo-spatial Information Science*, pp. 1–19, 2024.

[17] D. Peng, L. Bruzzone, Y. Zhang, H. Guan, H. Ding, and X. Huang, "SemiCDNet: A semisupervised convolutional neural network for change detection in high resolution remote-sensing images," *IEEE Transactions on Geoscience and Remote Sensing*, vol. 59, no. 7, pp. 5891–5906, 2020.

[18] S. Singh and R. Talwar, "A comparative study on change vector analysis based change detection techniques," *Sadhana*, vol. 39, pp. 1311–1331, 2014.